%% file: main.tex
\documentclass{article}
\usepackage{spconf,amsmath,epsfig}

\usepackage{bm}
\usepackage{amsfonts}
\usepackage{graphicx}
\usepackage{booktabs}
\usepackage{subfigure}
\usepackage{multirow}
\usepackage{algorithm}
\usepackage{algorithmic}

\usepackage{cite}
\usepackage{verbatim}
\usepackage{url}

\begin{document}\sloppy

\def\x{{\mathbf x}}
\def\L{{\cal L}}

\title{Improving Captioning for Low-Resource Languages by Cycle Consistency}
%
\name{Yike Wu\textsuperscript{1*}\thanks{$^{*}$Work performed while interning at IBM Research - China.}, Shiwan Zhao\textsuperscript{2$\dagger$}\thanks{$^\dagger$Corresponding author.}, Jia Chen\textsuperscript{3}, Ying Zhang\textsuperscript{1}, Xiaojie Yuan\textsuperscript{1}, Zhong Su\textsuperscript{2}}

\address{\textsuperscript{1}Nankai University\quad\quad \textsuperscript{2}IBM Research - China\quad\quad \textsuperscript{3}Carnegie Mellon University\\
wuyike@dbis.nankai.edu.cn, zhaosw@cn.ibm.com, jiac@cs.cmu.edu\\
\{zhangying, yuanxiaojie\}@dbis.nankai.edu.cn, suzhong@cn.ibm.com}

\maketitle

\begin{abstract}
Improving the captioning performance on low-resource languages by leveraging English caption datasets has received increasing research interest in recent years. 
Existing works mainly fall into two categories: translation-based and alignment-based approaches. 
In this paper, we propose to combine the merits of both approaches in one unified architecture. 
Specifically, we use a pre-trained English caption model to generate high-quality English captions, and then take both the image and generated English captions to generate low-resource language captions. 
We improve the captioning performance by adding the cycle consistency constraint on the cycle of image regions, English words, and low-resource language words.
Moreover, our architecture has a flexible design which enables it to benefit from large monolingual English caption datasets. 
Experimental results demonstrate that our approach outperforms the state-of-the-art methods on common evaluation metrics. 
The attention visualization also shows that the proposed approach really improves the fine-grained alignment between words and image regions.
\end{abstract}
\begin{keywords}
image captioning, low-resource language, cycle consistency, fine-grained alignment
\end{keywords}

\begin{figure*}[htbp] 
\centering 
\includegraphics[scale=0.35]{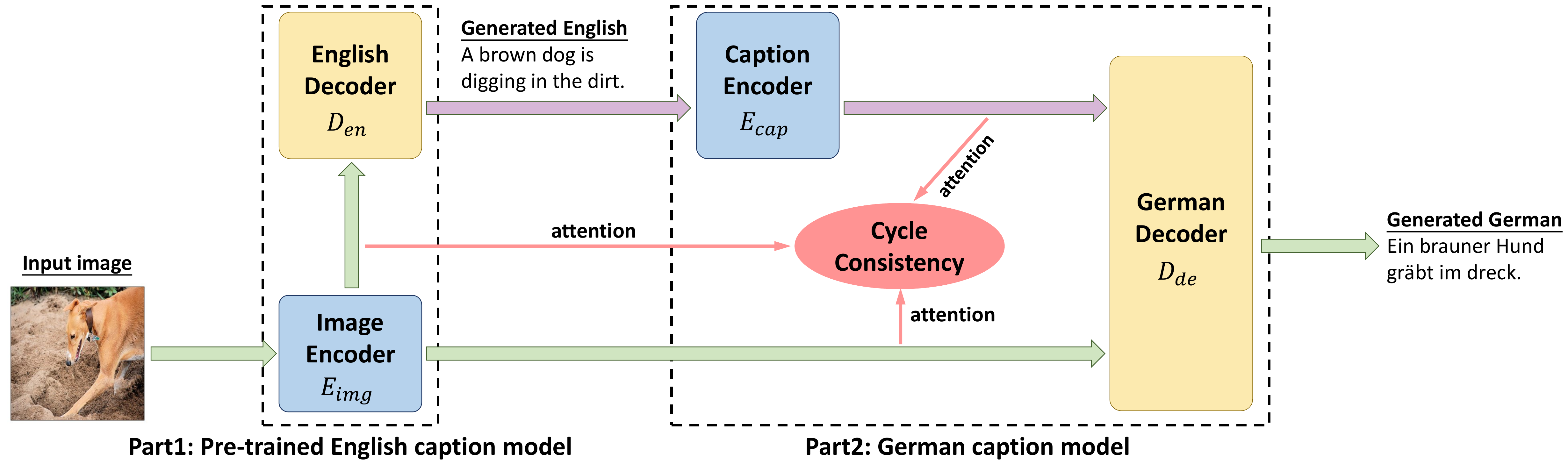}
\caption{Architecture overview. The thick arrows indicate the data flow of inference. The purple ones depict the translation process while the red thin arrows depict the fine-grained alignment among image regions, English words, and German words.} 
\label{model architecture}
\end{figure*}

\section{Introduction}
\input{introduction.tex}

\section{Methodology}
 We first provide an overview of the proposed architecture and then introduce each component in detail. Finally, the loss function and training process will be elaborated.

\subsection{Overview}
Fig.\ref{model architecture} shows an overview of the proposed architecture, which consists of two parts. \textbf{Part1} is a pre-trained English caption model, including an image encoder $E_{img}$ and an English decoder $D_{en}$. \textbf{Part2} is a German caption model, including an encoder $E_{cap}$ for generated English captions, a German decoder $D_{de}$, and a cycle consistency constraint. In the inference phase, we first feed an image into $E_{img}$ to get its corresponding English caption from $D_{en}$ as a pseudo English caption. Next, we feed the pseudo English caption into $E_{cap}$. Finally, $D_{de}$ generates a German caption by taking the outputs from both $E_{img}$ and $E_{cap}$.

\subsection{Pre-trained English Caption Model}
\label{softattn model}
For the English caption model in Part1, we follow the soft-attention approach proposed by \cite{sattn}. We use a pre-trained ResNet-152\cite{resnet} as the encoder $E_{img}$ and an LSTM\cite{lstm} as the decoder $D_{en}$. For a given image $I$, we feed it into $E_{img}$ to extract the feature vectors $\bm{V}=[\bm{v}_1;\bm{v}_2;...;\bm{v}_L]$. Then we calculate the attention weights $\bm{\alpha}_{t}^{en}$ and context vector $\bm{c}_t^{en}$ at every step $t$.
\begin{gather}
e_{ti}=f_{att}(\bm{v}_i, \bm{h}_{t-1}),\\
\alpha_{ti}^{en}=\frac{exp(e_{ti})}{\sum_{k=1}^Lexp(e_{tk})}, i\in\{1,2,...,L\},\\
\bm{c}_t^{en} = \phi(\bm{\alpha}_{t}^{en}, \bm{V}),
\end{gather}
where $f_{att}$ is an attention model based on the multilayer perceptron, $\bm{h}_{t-1}$ is the previous hidden state of $D_{en}$ and $\phi$ is a function that returns a weighted summation of the feature vectors $\bm{V}$ based on $\bm{\alpha}_t^{en}$. At last, the output word probability is calculated conditioned on $\bm{c}_t^{en}$, $\bm{h}_{t-1}$ and the embedding of the previously generated English word $y_{t-1}^{en}$:
\begin{equation}
\begin{split}
    p(y_t^{en}|\bm{c}_t^{en}, \bm{h}_{t-1},\bm{y}_{t-1}^{en}) = &\\ \mathrm{Softmax}(\mathrm{LSTM}&([\bm{c}_t^{en};\bm{y}_{t-1}^{en}], \bm{h}_{t-1})),
\end{split} 
\end{equation}
here we use the same notation for the word and its embedding with a slight abuse of notations.

\subsection{German Caption Model}
For the model in Part2, we use a bidirectional GRU\cite{gru} as the English caption encoder $E_{cap}$ and follow the doubly-attentive architecture \cite{dattn, multilingual2} for the German decoder $D_{de}$. The German decoder $D_{de}$ has two attentions over image regions and English words respectively. For the former, we compute the context vector $\bm{c}_t^{de}$ in a similar way of $\bm{c}_t^{en}$. For the latter, we calculate the attention weights $\bm{\beta}_t$ over the hidden states $\bm{G}=[\bm{g}_1;\bm{g}_2;...;\bm{g}_N]$ of $E_{cap}$ and the context vector $\bm{z}_t$ at every step $t$ as follows:
\begin{gather}
e'_{tj}=f'_{att}(\bm{g}_j, \bm{s}_{t-1}),\\
 \beta_{tj}=\frac{exp(e'_{tj})}{\sum_{k=1}^Nexp(e'_{tk})}, j\in\{1,2,...,N\},\\
 \bm{z}_t = \phi(\bm{\beta}_t, \bm{G}),
\end{gather}
where $f_{att}'$ is the attention model, and $\bm{s}_{t-1}$ is the previous hidden state of $D_{de}$. Thus the output word probability is computed as follows:
\begin{equation}
\begin{split}
    p(y_t^{de}|\bm{c}_t^{de},\bm{z}_t,\bm{s}_{t-1},\bm{y}_{t-1}^{de}) = &\\ \mathrm{Softmax}(\mathrm{LSTM}&([\bm{c}_t^{de};\bm{z}_t;\bm{y}_{t-1}^{de}], \bm{s}_{t-1})).
\end{split}  
\end{equation}

\subsection{Cycle Consistency}
\begin{figure}[htbp] 
\centering 
\includegraphics[height=5cm, width=7.5cm]{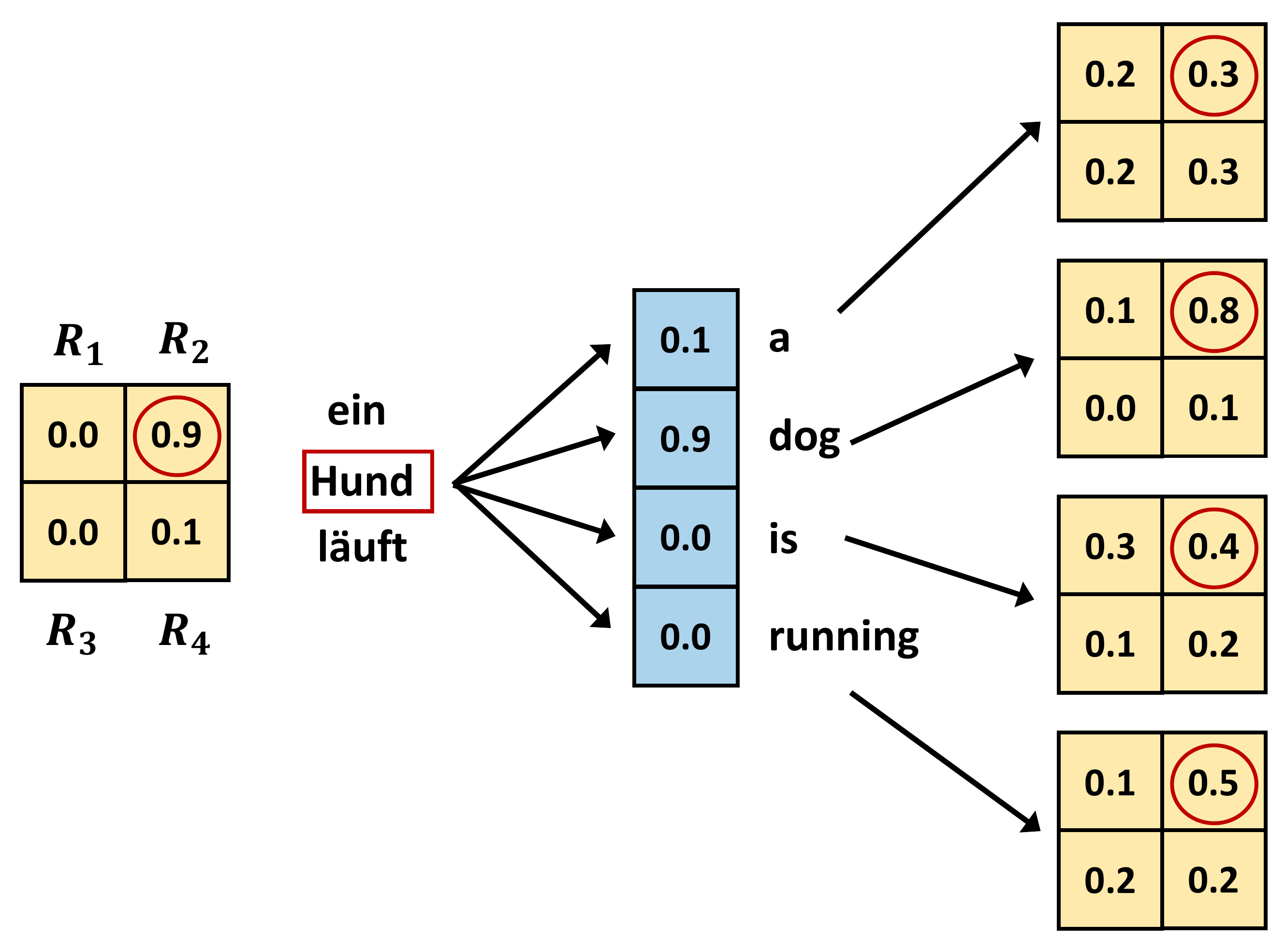}
\caption{A toy example of the cycle consistency. The direct attention of \emph{ ``Hund''} on $R_2$ is $0.9$. And the indirect attention of \emph{ ``Hund''} on $R_2$ is $0.75$ $(0.1\times0.3+0.9\times0.8+0.0\times0.4+0.0\times0.5=0.75)$.} 
\label{cycle consitency}
\end{figure}

Fig.\ref{cycle consitency} shows a toy example of the cycle consistency of image regions, English words, and German words. We assume that the attention weights of \emph{``Hund''} over the English words are $(0.1, 0.9, 0.0, 0.0)$ in the word order of the sentence. For simplicity, we assume that the image only has four regions $\{R_1,R_2,R_3,R_4\}$, and the attention weights of \emph{``Hund''} over these regions are $(0.0,0.9,0.0,0.1)$. Similarly, each word in the English caption has a set of attention weights over these regions. If we want to know the attention weight of \emph{``Hund''} on $R_2$, there are two ways to get the answer. One is to pick it out directly, and the value is 0.9. We call it the direct attention. The other is to sum the attention weights on $R_2$ of every word in the English caption based on their relative importance on the generation of \emph{``Hund''}. As Fig.\ref{cycle consitency} shows, the value calculated in this way is 0.75, and we call it the indirect attention. Theoretically, the values got in these two ways should be equal if the attentions are computed accurately. This is exactly the cycle consistency. Moreover, for a caption model, the more accurate the attention is, the more reasonable captions it generates. Therefore, it is natural to improve the quality of the low-resource language captioning by guaranteeing the cycle consistency. 

Now we describe the cycle consistency in a formal way. Given an {\tt Image-English-German} triple, each word $y_m^{de}$ in the German caption has two sets of attention weights $\bm{\alpha}_m^{de}$ and $\bm{\beta}_m$ over the image and English caption respectively. In addition, each word $y_j^{en}$ in the English caption has a set of attention weights $\bm{\alpha}_j^{en}$ over the image. If the models in our architecture were all perfect, the following equation should be established:
\begin{equation}
\begin{split}
    \alpha_{mi}^{de}&=\sum_{j=1}^N\beta_{mj}\alpha_{ji}^{en},\\
    &i\in\{1,2,...,L\},m\in\{1,2,...,M\},\label{cycle consistency}
\end{split}
\end{equation}
where M is the length of the German caption, and N is the length of the English caption.


Next we briefly explain why Eq.\ref{cycle consistency} is mathematically correct. We consider a set of attention weights as a conditional probability distribution,
image regions as random variable $X$, English words $\{Y_1,Y_2,...,Y_j,...,Y_N\}$ as random variable $Y$, and German words as random variable $Z$. Obviously, $X$ and $Z$ are conditional independent given $Y$. Then Eq.\ref{cycle consistency} can be written as:
\begin{equation}
\begin{split}
P(X|Z)=\sum_{j=1}^NP(X,Y_j|Z)= & \sum_{j=1}^NP(X|Y_j,Z)P(Y_j|Z)\\
= & \sum_{j=1}^NP(X|Y_j)P(Y_j|Z).
\label{prob}
\end{split}
\end{equation}

\subsection{Loss Function}
 The Loss function of the proposed approach is composed of two parts. One is the summation of negative log likelihood of the German word (superscript omitted here) at each step $t$:
\begin{equation}
    \mathcal{L}_{nll} = -\sum_{t=1}^{M}\log p(y_{t}|y_1,y_2,...,y_{t-1}).
\end{equation}
The other is the cycle consistency loss, the Euclidean Distance between the direct attention and indirect attention:
\begin{equation}
    \mathcal{L}_{cyc} = ||\bm{A}^{de} - \bm{B}\bm{A}^{en}||_2,
\end{equation}
where $\bm{A}^{de}=[\bm{\alpha}_1^{de};\bm{\alpha}_2^{de};...;\bm{\alpha}_M^{de}]^\mathrm{T}$ is a matrix of size $M \times L$, $\bm{B}=[\bm{\beta}_1;\bm{\beta}_2;...;\bm{\beta}_M]^\mathrm{T}$ is a matrix of size $M \times N$, and $\bm{A}^{en}=[\bm{\alpha}_1^{en};\bm{\alpha}_2^{en};...;\bm{\alpha}_N^{en}]^\mathrm{T}$ is a matrix of size $N \times L$.

\subsection{Training Process}

We elaborate the training process in Algorithm \ref{optim}, which can be divided into three stages. At the first stage, we pre-train the English caption model in Part1 using {\tt Image-English} pairs. At the second stage, we train the German caption model in Part2 using {\tt Image-English-German} triples. Specifically, we infer the two attentions of German words over image regions and English words, and calculate $\mathcal{L}_{nll}$. Then, to form the attention cycle, we further extract Image-English pairs from the {\tt Image-English-German} triples, and feed them into the pre-trained English caption model to get the attention of English words over image regions. Finally, with these three attentions, we then compute $\mathcal{L}_{cyc}$.   
At the third stage, we update model parameters with $\mathcal{L}_{nll}$ and $\mathcal{L}_{cyc}$. 

 It is worth noting that the {\tt Image-English} pairs for Part1 may come from the {\tt Image-English-German} triples used in Part2, or any other large monolingual dataset.


\begin{algorithm}[htb]
\caption{Training Process}
 \label{optim}
\begin{algorithmic}[1]
\REQUIRE~~\\
{\tt Image-English} pairs, \\ mini-batches of {\tt Image-English-German} triples $\{b_1, b_2,...,b_n\}$,\\ randomly initialized models $E_{img}$, $E_{cap}$, $D_{en}$, $D_{de}$, and their parameters $\Theta$
\ENSURE~~\\
Trained model parameters $\Theta$.
\STATE{pre-train $E_{img}$ and $D_{en}$ using {\tt Image-English} pairs}
\WHILE{not converge}
\FORALL{$b$ in $\{b_1, b_2,...,b_n\}$}
\STATE{infer $\alpha^{de}$ to align $E_{img}$ and $D_{de}$}
\STATE{infer $\beta$ to align $E_{cap}$ and $D_{de}$}
\STATE{infer $\alpha^{en}$ to align $E_{img}$ and $D_{en}$}
\STATE{calculate $\mathcal{L}_{cyc}$ for $\alpha^{de}$, $\beta$, $\alpha^{en}$}
\STATE{calculate $\mathcal{L}_{nll}$ for $D_{de}$}
\STATE{update $\Theta$ with $\nabla\mathcal{L}_{nll}+\nabla\mathcal{L}_{cyc}$}
\ENDFOR
\ENDWHILE
\RETURN $\Theta$
\end{algorithmic}
\end{algorithm}

\section{EXPERIMENTS}
In this section, we first introduce the dataset and experimental settings. Then, we compare our approach with the baselines on common metrics. Finally, we validate the effectiveness of cycle consistency on fine-grained alignment by visualizing the attentions.

\subsection{Dataset}
The Flickr30K dataset\cite{flickr} consists of 29k, 1,014 and 1k images for training, validation and testing respectively. Each image is associated with five English captions. The Multi30K dataset extends Flickr30K in two ways with translated and independent German sentences. To form a cycle by combining both translation-based and alignment-based approaches, we perform experiments on the translated version of Multi30K, denoted by {\bf Multi30K-Trans}. For each image in Flickr30K, Multi30K-Trans adds a manually translated German caption for only one of the English captions to compose an {\tt Image-English-German} triple. 


\subsection{Experimental Settings}
\textbf{Data Preprocessing} Images are resized to $450\times450$ for uniformity and then fed into ResNet-152 to extract features using the layer before the penultimate average pooling layer. We don't finetune ResNet-152 considering the time cost. When building English and German vocabularies, we remove punctuations and filter the tokens whose frequency is below 5.
    
\textbf{Model and Training} The hidden size of LSTM and embedding size are 512, and dropout rate is 0.5 for all models. Maximum epoch is set to 50 and we apply early stopping for model selection if a model does not improve the performance on the validation set on CIDEr for more than 20 epochs. And we use Adam optimizer\cite{adam} with a learning rate of $4\times10^{-4}$ and the batch size of 32.

\textbf{Inference and Evaluation} Beam size for inference is 3 and generated captions longer than 50 tokens are discarded. We evaluate the inference results on metrics CIDEr, BLEU4 and METEOR based on the provided implementation\footnote{\url{https://github.com/tylin/coco-caption}}.

\begin{figure*}[htbp]
    \centering
    \includegraphics[scale=0.45]{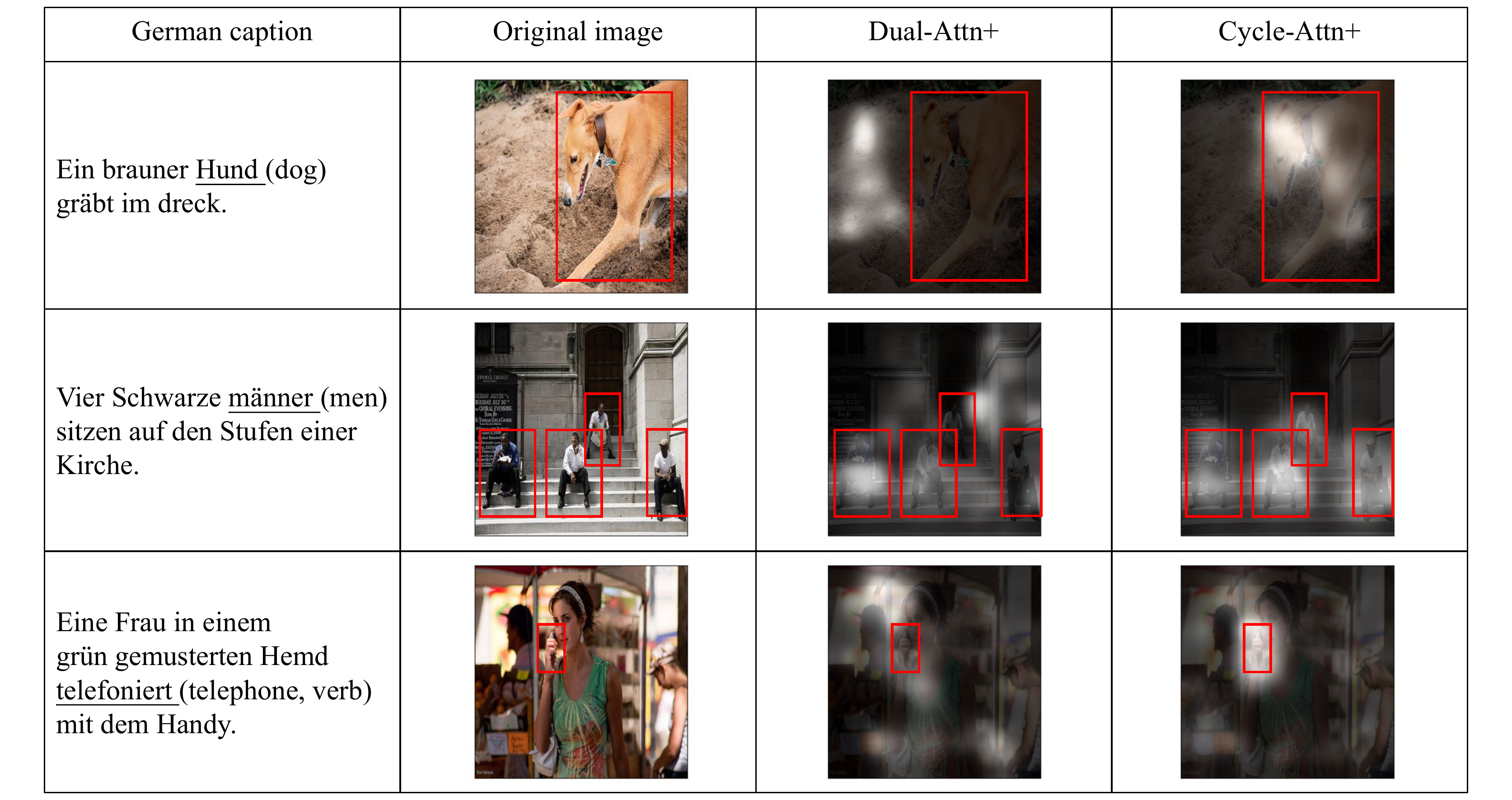}
    \caption{Attention visualization on German captions. Follow the style in \cite{sattn}, \textit{white} indicates the attended regions, and \textit{underlines} indicate the corresponding words. For better readability, we display the English definitions of German words in brackets, and identify the regions which the attention should focus on with red frames.}
    \label{visual_attn}
    \vspace{-10pt}
\end{figure*}

\subsection{Quantitative Analysis}
We first perform experiments on Multi30K-Trans to validate the effectiveness of the cycle consistency of our approach. The {\tt Image-Englsh} pairs for Part1 are extracted from the {\tt Image-English-German} triples of Multi30K-Trans. We denote our approach as \textbf{Cycle-Attn} and compare it with the following baselines:
\begin{itemize}
    \item \textbf{Trans}\cite{addchinese} This method first pre-trains a machine translation model, then translates generated English captions into German directly.
    \item \textbf{Soft-Attn}\cite{sattn} It trains a soft attention caption model on images and corresponding German captions directly.
    \item \textbf{Dual-Attn}\cite{multilingual2} It trains an English caption model and a doubly-attentive model for generating German captions. When testing, it uses generated English captions from the pre-trained model as pseudo English captions.
\end{itemize}


Moreover, to demonstrate that our architecture can benefit from large monolingual English caption datasets, we use the {\tt Image-English} pairs from Flickr30K, which has more (five) English captions for each image, to pre-train the Part1 English caption model. We denote this variant of Cycle-Attn as \textbf{Cycle-Attn+}. In addition, we also provide a variant of Dual-Attn (denoted by \textbf{Dual-Attn+}) for fair comparison. 
 
 \begin{table}[htbp]
    \centering
    \begin{tabular}{lllll}
    \toprule
    Model  &  CIDEr &  BLEU4 &  METEOR\\
    \midrule
      Trans\cite{addchinese} & 37.82 & 5.28 & 10.27\\
      Soft-Attn\cite{sattn} & 38.59 & 5.12 &  \textbf{10.86}\\
      Dual-Attn\cite{multilingual2} & 40.57 &  5.32 &  10.51\\
      Cycle-Attn & 41.91 &  5.67 &  10.59\\
    \midrule
    \midrule
      Dual-Attn+ & 42.91 & 5.54 & 10.79\\
      Cycle-Attn+ & \textbf{43.78} & \textbf{5.71} &\textbf{10.86}\\ 
    \bottomrule
    \end{tabular}
    \caption{Experimental results on common metrics.}
    \label{main results}
\end{table}
 
Table \ref{main results} shows the experimental results. Firstly, we observe that Cycle-Attn outperforms both Soft-Attn and Trans in most metrics. Particularly, Cycle-Attn outperforms Soft-Attn by $+3.32\ (8.60\%)$ CIDEr, and $+0.55\ (10.74\%)$ BLEU4. This observation demonstrates that the proposed architecture can improve low-resource language captioning of either translation-based or alignment-based approach by combining their merits. Secondly, Cycle-Attn achieves better performance on all metrics comparing with Dual-Attn. For example, it improves CIDEr by $+1.34\ (3.30\%)$ and BLEU4 by $+0.35\ (6.58\%)$, which validates the effectiveness of the cycle consistency. Finally, by pre-training the English caption model on Flickr30K instead of Multi30K-Trans, Cycle-Attn+ outperforms Cycle-Attn on all metrics. This demonstrates the notable benefit from rich-resource datasets for the proposed architecture. Moreover, Cycle-Attn+ also performs better than Dual-Attn+ on all metrics. This indicates that our approach can benefit from the cycle consistency and rich-resource dataset simultaneously.

\subsection{Qualitative Analysis}
We visualize the attention weights obtained by Dual-Attn+ and Cycle-Attn+ in Fig.\ref{visual_attn}. Specifically, we feed the same image and its German ground truth into both models to infer the attention weights over the image. Note that we use the German ground truth here because generated German captions from the two models may contain different words, which is not conducive to fair comparison. As we can see in Fig.\ref{visual_attn}, there are three images, each of which is a representative situation that the attention mechanism needs to handle. The first and second rows represent a single-object situation and a multiple-object situation respectively, and the third row focuses on the detail of an image which is hard to capture. 

Now we compare the quality of the attentions. We observe that Cycle-Attn+ performs better than Dual-Attn+ in all situations significantly. Particularly, in the multiple-object situation, Cycle-Attn+ even outlines all four people in the image. This fully demonstrates that the cycle consistency really helps the model learn better fine-grained alignment, which leads to better German captions.

\section{CONCLUSION}
In this paper, we propose a method to combine the merits of existing approaches to improve low-resource language captioning in one unified architecture. 
The proposed method incorporates generated English captions into generating low-resource language captions, and improve the fine-grained alignment by cycle consistency. 
Flexible architecture of the proposed method also enables us to benefit from large monolingual English caption datasets. 
Experimental results demonstrate that the proposed method really achieves better performance on common evaluation metrics comparing with the state-of-the-art methods and improves the fine-grained alignment. 
In the future, we plan to improve image captioning for low-resource languages distant from English, such as Japanese, which are difficult to align with English in the joint embedding space.

\textbf{ACKNOWLEDGEMENTS} This research is supported by National Natural Science Foundation of China (No. U1836109) and Tianjin Natural Science Foundation (NO. 18ZXZNGX00110).

\bibliographystyle{IEEEbib}
\bibliography{myreferences_short}

\end{document}

%% file: introduction.tex
Automatically generating image captions is an important and challenging task in the intersection between computer vision and natural language processing. 
Recent years have witnessed exciting progress in this field based on deep learning methods \cite{nic, sattn, self-critical, congan, nbtalk, topdown}. Most caption datasets\cite{flickr8k, flickr, coco} in these works are collected in the English language. 
However, for people who don't speak English, there are strong needs for image captioning in languages other than English. 
There are some caption datasets\cite{multi30k, crosslingual} collected in languages other than English, but the scale of these datasets is relatively small compared to that of various English caption datasets. 
Thus, such languages are considered as a low-resource language for the captioning task. Improving the captioning performance on low-resource languages by leveraging English caption datasets has received increasing research interest in recent years. 

To improve the captioning performance on a low-resource language with an English caption dataset, current works \cite{addchinese, fluency, unpaired, crosslingual, multilingual} can be categorized into two different approaches: translation-based and alignment-based. 
The first kind of approach is based on translation\cite{addchinese, fluency, unpaired}. 
Based on machine translation models, they usually translate generated English captions into the low-resource language, or exploit these translations to construct a pseudo caption corpus and train a caption model for the low-resource language. 
However, these methods are limited by the quality of translations and suffer from the difference of data distributions between caption data and translation data. 
The second kind of approach is based on alignment in the joint embedding space.
The rationale of these methods is to learn better alignment between images and their corresponding sentences in a common latent space by involving English captions, and better alignment leads to better quality of caption generation in the low-resource language. Miyazaki and Shimizu \cite{crosslingual} enhance the encoder of a Japanese caption model by pre-training it on a large English caption dataset MSCOCO\cite{coco}.  Elliott \emph{et al.} \cite{multilingual} propose a multimodal architecture to generate captions from the features of both images and English captions. 
These models actually do coarse-level alignment between images and sentences in the joint embedding space. 

In this work, we propose to combine the merits of both approaches in one unified architecture. 
To be specific, we design an architecture which first generates English captions from the image and then generates low-resource language (\emph{i.e.}, German) captions given both the image and the generated English captions as shown in Fig.~\ref{model architecture}.
There are three advantages of the proposed architecture. 
First, the English decoder could benefit from rich-resource English caption datasets through pre-training. 
As the English decoder is decoupled from other parts in our architecture, we could pre-train the English decoder on a large monolingual English caption dataset and then finetune it with other parts in the architecture on a multilingual dataset. 
Second, the low-resource language decoder benefits from the generated high-quality English captions. 
In our architecture, the low-resource language decoder depends not only on the image but also on the generated English captions. 
The dependency on the English captions could be considered as imitating the translation-based approach. 
Third, we introduce fine-grained alignment between image regions, English words and low-resource language words through cycle consistency. 
For example, the image region of a dog should correspond to the word \emph{``dog''} in the English caption and word \emph{``Hund''} in the German caption simultaneously. 
We achieve this by adding cycle consistency on three attentions: 
the attention in the English decoder conditioned on image regions, 
the attention in the low-resource language decoder conditioned on image regions, and another attention in the low-resource language decoder conditioned on English words. 
These three attentions should be consistent in the cycle of image regions, English words and low-resource language words. 

In summary, our main contributions are as follows:
\begin{itemize}
    \item We propose an architecture that combines the merits of both translation-based and alignment-based approaches to improve low-resource language captioning. 
    \item We improve the performance by adding the cycle consistency constraint on the cycle of image regions, English words and low-resource language words. 
    \item Our architecture has a flexible design which enables it to benefit from large monolingual English caption datasets. 
\end{itemize}

%% file: main.bbl
\begin{thebibliography}{10}

\bibitem{nic}
Oriol Vinyals, Alexander Toshev, Samy Bengio, and Dumitru Erhan,
\newblock ``Show and tell: A neural image caption generator,''
\newblock in {\em CVPR}, 2015.

\bibitem{sattn}
Kelvin Xu, Jimmy Ba, Ryan Kiros, Kyunghyun Cho, Aaron~C. Courville, Ruslan
  Salakhutdinov, Richard~S. Zemel, and Yoshua Bengio,
\newblock ``Show, attend and tell: Neural image caption generation with visual
  attention,''
\newblock in {\em ICML}, 2015.

\bibitem{self-critical}
Steven~J. Rennie, Etienne Marcheret, Youssef Mroueh, Jarret Ross, and Vaibhava
  Goel,
\newblock ``Self-critical sequence training for image captioning,''
\newblock in {\em CVPR}, 2017.

\bibitem{congan}
Bo~Dai, Dahua Lin, Raquel Urtasun, and Sanja Fidler,
\newblock ``Towards diverse and natural image descriptions via a conditional
  gan,''
\newblock in {\em ICCV}, 2017.

\bibitem{nbtalk}
Jiasen Lu, Jianwei Yang, Dhruv Batra, and Devi Parikh,
\newblock ``Neural baby talk,''
\newblock in {\em CVPR}, 2018.

\bibitem{topdown}
Peter Anderson, Xiaodong He, Chris Buehler, Damien Teney, Mark Johnson, Stephen
  Gould, and Lei Zhang,
\newblock ``Bottom-up and top-down attention for image captioning and visual
  question answering,''
\newblock in {\em CVPR}, 2018.

\bibitem{flickr8k}
Micah Hodosh, Peter Young, and Julia Hockenmaier,
\newblock ``Framing image description as a ranking task: Data, models and
  evaluation metrics (extended abstract),''
\newblock {\em J. Artif. Intell. Res.}, vol. 47, pp. 853--899, 2013.

\bibitem{flickr}
Peter Young, Alice Lai, Micah Hodosh, and Julia Hockenmaier,
\newblock ``From image descriptions to visual denotations: New similarity
  metrics for semantic inference over event descriptions,''
\newblock {\em TACL}, vol. 2, pp. 67--78, 2014.

\bibitem{coco}
Tsung-Yi Lin, Michael Maire, Serge~J. Belongie, Lubomir~D. Bourdev, Ross~B.
  Girshick, James Hays, Pietro Perona, Deva Ramanan, Piotr Doll{\'a}r, and
  C.~Lawrence Zitnick,
\newblock ``Microsoft coco: Common objects in context,''
\newblock in {\em ECCV}, 2014.

\bibitem{multi30k}
Desmond Elliott, Stella Frank, Khalil Sima'an, and Lucia Specia,
\newblock ``Multi30k: Multilingual english-german image descriptions,''
\newblock {\em CoRR}, vol. abs/1605.00459, 2016.

\bibitem{crosslingual}
Takashi Miyazaki and Nobuyuki Shimizu,
\newblock ``Cross-lingual image caption generation,''
\newblock in {\em ACL}, 2016.

\bibitem{addchinese}
Xirong Li, Weiyu Lan, Jianfeng Dong, and Hailong Liu,
\newblock ``Adding chinese captions to images,''
\newblock in {\em ICMR}, 2016.

\bibitem{fluency}
Weiyu Lan, Xirong Li, and Jianfeng Dong,
\newblock ``Fluency-guided cross-lingual image captioning,''
\newblock in {\em ACM Multimedia}, 2017.

\bibitem{unpaired}
Jiuxiang Gu, Shafiq~R. Joty, Jianfei Cai, and Gang Wang,
\newblock ``Unpaired image captioning by language pivoting,''
\newblock in {\em ECCV}, 2018.

\bibitem{multilingual}
Desmond Elliott, Stella Frank, and Eva Hasler,
\newblock ``Multilingual image description with neural sequence models,''
\newblock {\em arXiv preprint arXiv:1510.04709}, 2015.

\bibitem{resnet}
Kaiming He, Xiangyu Zhang, Shaoqing Ren, and Jian Sun,
\newblock ``Deep residual learning for image recognition,''
\newblock in {\em CVPR}, 2016.

\bibitem{lstm}
Sepp Hochreiter and J{\"u}rgen Schmidhuber,
\newblock ``Long short-term memory,''
\newblock {\em Neural Computation}, vol. 9, pp. 1735--1780, 1997.

\bibitem{gru}
Kyunghyun Cho, Bart van Merrienboer, Dzmitry Bahdanau, and Yoshua Bengio,
\newblock ``On the properties of neural machine translation: Encoder-decoder
  approaches,''
\newblock in {\em SSST@EMNLP}, 2014.

\bibitem{dattn}
Iacer Calixto, Qun Liu, and Nick Campbell,
\newblock ``Doubly-attentive decoder for multi-modal neural machine
  translation,''
\newblock in {\em ACL}, 2017.

\bibitem{multilingual2}
Alan Jaffe,
\newblock ``Generating multilingual2image descriptions using multilingual
  data,''
\newblock in {\em WMT}, 2017.

\bibitem{adam}
Diederik~P. Kingma and Jimmy Ba,
\newblock ``Adam: A method for stochastic optimization,''
\newblock {\em CoRR}, vol. abs/1412.6980, 2014.

\end{thebibliography}
